\documentclass[journal]{IEEEtran}
\ifCLASSINFOpdf
\else
\fi
\usepackage[dvipsnames]{xcolor}
\usepackage{graphicx}
\usepackage{tikz}
\usepackage{graphicx}
\usepackage{listings}
\usepackage{amsmath}
\usepackage{commath}
\usepackage{pgfplots}
\usepackage{colortbl}
\usepackage{algorithmicx}
\usepackage{algorithm}   
\usepackage{caption}
\usepackage{subcaption}
\usepackage[noend]{algpseudocode}
\algrenewcommand\algorithmicindent{0.8em}%
\usetikzlibrary{arrows, shapes.arrows, decorations.pathreplacing, calligraphy, fit}
\usetikzlibrary{positioning}
\usetikzlibrary{shapes.geometric, arrows}
\tikzstyle{state} = [rectangle, text width=1.5cm, minimum height=1.5cm, rounded corners, text centered, draw=black, fill=gray!10]
\tikzstyle{state1} = [rectangle, text width=2.1cm, minimum height=1.5cm, rounded corners, text centered, draw=black, fill=gray!10]
\tikzstyle{stage} = [rectangle, text width=1.2cm, rounded corners, text centered]
\tikzstyle{stage2} = [rectangle, text width=1cm, rounded corners, text centered]
\tikzstyle{stage3} = [rectangle, minimum height=0.8cm, text width=0.2cm, rounded corners, text centered]
\tikzstyle{arrow} = [->,>=stealth]
\tikzstyle{th_arrow} = [->, ultra thick]
\tikzstyle{conv} = [trapezium, trapezium angle=60, minimum width=1cm, draw, thick, rounded corners, rotate=270]

\definecolor{codegreen}{rgb}{0,0.6,0}
\definecolor{codegray}{rgb}{0.5,0.5,0.5}
\definecolor{codepurple}{rgb}{0.58,0,0.82}
\definecolor{backcolour}{rgb}{0.95,0.95,0.92}

\lstdefinestyle{mystyle}{
    backgroundcolor=\color{backcolour},   
    commentstyle=\color{codegreen},
    keywordstyle=\color{magenta},
    numberstyle=\tiny\color{codegray},
    stringstyle=\color{codepurple},
    basicstyle=\ttfamily\footnotesize,
    breakatwhitespace=false,         
    breaklines=true,                 
    captionpos=b,                    
    keepspaces=true,                 
    numbers=left,                    
    numbersep=5pt,                  
    showspaces=false,                
    showstringspaces=false,
    showtabs=false,                  
    tabsize=2,
    morekeywords={split, vectorize, parallel, const, int, reorder, realize, update, Var, compute_at, ImageParam, float, type_of, Func, RDom, RVar, Buffer, compute_root},
}

\lstdefinestyle{mystyle2}{
    basicstyle=\footnotesize,
    backgroundcolor=\color{white},
    breakatwhitespace=false,         
    breaklines=true,                 
    captionpos=b,                    
    keepspaces=true,                 
    numbers=left,                    
    numbersep=5pt,                  
    showspaces=false,                
    showstringspaces=false,
    showtabs=false,                  
    tabsize=2,
    morekeywords={produce, for, consume},
}

\begin{document}
%
\title{Using Graph Neural Networks to model the performance of Deep Neural Networks}

\author{Shikhar~Singh,
        James~Hegarty,
        Hugh~Leather,
        Benoit~Steiner
\thanks{Shikhar Singh is with the Department of Electrical and Computer Engineering, University of Texas, Austin at the time of publication and was an intern at Facebook at the time of this work.}
\thanks{Benoit Steiner, James Hegarty, and Hugh Leather are with Facebook, Inc.}
\thanks{Manuscript received August 16, 2021; revised August 20, 2021.}
}

\markboth{}
{Shell \MakeLowercase{\textit{et al.}}: Bare Demo of IEEEtran.cls for IEEE Journals}
%



\maketitle

\begin{abstract}
The unprecedented proliferation of machine learning based software brings an ever-increasing need to optimize the implementation of such applications. State-of-the-art compilers for neural networks, such as Halide or TVM, incorporate a machine learning-based performance model to search the space of valid implementations of a given deep learning algorithm. For a given application, the model predicts the value of performance metrics such as the run time without executing the application on hardware. Such models speed up the compilation process by obviating the need to benchmark an enormous number of candidate implementations, referred to as \emph{schedules}, on hardware. Existing performance models employ feed-forward networks, recurrent networks, or decision tree ensembles to estimate the performance of different implementations of a neural network. 
Graphs present a natural and intuitive way to model deep-learning networks where each node represents a computational stage or operation.
Incorporating the inherent graph structure of these workloads in the performance model can enable a better representation and learning of inter-stage interactions. 
The accuracy of the performance model has direct implications on the efficiency of the search strategy, making it a crucial component of this class of deep-learning compilers. 
In this work, we develop a novel performance model that adopts a graph representation. In our model, each stage of computation represents a node characterized by features that capture the operations performed by the stage. The interaction between nodes is achieved using graph convolutions. Experimental evaluation shows a $7.75x$ and $12x$ reduction in prediction error compared to the Halide and TVM models, respectively.

\end{abstract}


%
\IEEEpeerreviewmaketitle

\section{Introduction}\label{intro}
Compilation frameworks like Halide~\cite{halide} and TVM~\cite{tvm} use a high-level domain-specific language to express tensor computations. Deep learning programs are written in this high-level language, which gets translated into low-level optimized code for a particular hardware platform. This approach separates the algorithmic computation, also known as the \emph{pipeline}, from the way it is performed, called the \emph{schedule} - a technique first proposed in Halide. Users can specify the computations to realize their machine learning model, and the compiler generates an efficient low-level implementation for the model. Depending on the target hardware architecture, the compiler has to determine an appropriate set of optimizations for the efficient execution of the application. In contrast, frameworks like Tensorflow~\cite{tf} and PyTorch~\cite{pytorch} rely on a library of optimized low-level implementations of common tensor operations, such as matrix multiplication, pooling, etc. These implementations are hardware-specific and are expert-written. Neural networks are represented as a computational graph, and the compilation process translates each operation into its optimized low-level counterpart. This approach has two drawbacks. First, it requires a lot of human effort to create and maintain these libraries. The implementations have to be rewritten and revised for new hardware and iterations, respectively. Secondly, code optimization happens at the operator level-granularity and often overlooks critical graph-level optimizations.

For a given deep-learning computation (or any computation for that matter), the quality of a schedule (how the computation is performed) depends on how well it can utilize the available hardware resources. For example, a convolution operation on a CPU will be more efficient if it can exploit the underlying cache hierarchy and locality using tiling and reordering of loads/stores. The computation will also take less time if the schedule can take advantage of the data parallelism provided by the hardware via \emph{SIMD} or \emph{vector} instructions. While Halide and TVM, in principle, have the capability to explore such optimization avenues, they have their own set of challenges. For a given neural network computation, a compiler has to explore an enormous number of schedules in a space parameterized by per-stage choices like loop unrolling, vectorization, tiling, etc. Schedule search is a time consuming combinatorial problem since any practical neural network represents a vast schedule space. To make the search process more efficient, analytical models that guide the search towards efficient schedules have been proposed.

\begin{figure}[h]
  \centering
  \begin{tikzpicture}
    \node (a) [state] {\small Deep Learning Application};
    \node (b) [state, right of=a, xshift=1.2cm] {\small Feature Engineering};
    \node (c) [state, right of=b, xshift=1.2cm] {\small Model};
    \node (d) [state, right of=c, xshift=1.2cm] {\small Run Time};
    \node (e) [state, below of=d, yshift=-1cm, xshift=-3.2cm] {\small Benchmark};
    
    \draw [arrow] (a) -- (b);
    \draw [arrow] (b) -- (c);
    \draw [arrow] (c) -- (d);

    \draw [arrow, dashed] (a.south) |- (e.west);
    \draw [arrow, dashed] (e.east) [out=270, in=90] -| (d.south);
    \draw [arrow, dashed] (d.north) -- ++(0,1) node[anchor=south, xshift=-1.1cm] {\small{Train/Tune}} -| (c.north);
    
  \end{tikzpicture}
  \caption{\emph{Performance model construction:} Features obtained from a deep learning program are provided as inputs to the model. The output is the estimated run time of the program. The actual run-time of the application is used to train and tune the model.}
  \label{intro1}
\end{figure}
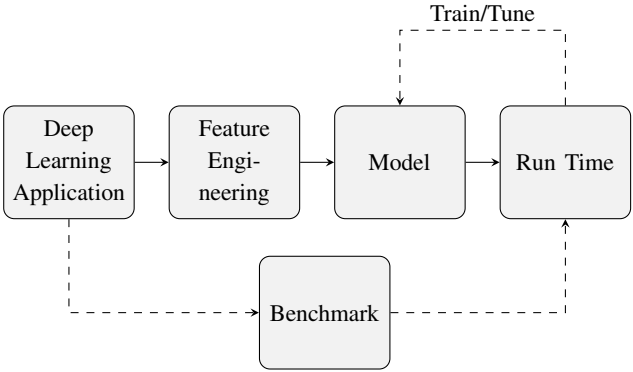

The main idea behind a model like the one we have developed is to estimate the performance of a given implementation without actually running it on hardware. 
Figure \ref{intro1} shows the typical construction methodology and working of a deep learning-based performance model like ours. 
The model takes as input features extracted from a deep-learning program. These features are designed to convey information about the structure of the network and the computation being performed. How these features are encoded depends on the model itself. The performance model uses these features and generates a performance metric, the run time in our case. Models are trained/tuned using the actual performance metrics obtained by benchmarking the input programs on the target hardware. 
Performance models offer a faster alternative to auto-tuning approaches when exploring schedules. This paper is another step towards improving the accuracy of the statistical performance models by utilizing the inherent graph structure of deep-learning networks. Specifically, we employ graph neural networks to estimate the performance of deep-learning pipelines in the Halide framework.

Previous models developed for the Halide framework used feed-forward~\cite{halide_auto} and recurrent neural network~\cite{value_learn} architectures. TVM~\cite{tvm_model} relies on  a decision tree based model. In this work, we introduce a novel performance model capable of utilizing the graph structure of deep-learning networks to capture inter-stage relationships. Incorporating neighborhood interactions using graphs can improve the quality of the  predictions. Our approach represents the pipeline as a directed acyclic graph (DAG) and aggregates information about the neighboring stages using graph convolutions~\cite{gcn}. Experimental evaluation of the proposed design demonstrates a $7.75x$ and $12x$ reduction in prediction error compared to the Halide and TVM models, respectively. 

This paper makes the following contributions.

\begin{itemize}
    \item{\textbf{A novel Graph Convolutional Network-based performance model.}} We propose a novel machine learning-based model to predict the execution time of deep learning pipelines. Our model can capture interactions between neighboring nodes to generate richer representations that improve the accuracy of the predictions.
    \item{\textbf{Implementation and Halide Interface.}} 
    We implement the proposed model in \emph{PyTorch}. The model takes as input features extracted from Halide pipelines and schedules and outputs the run time. To train the model, we generate a dataset comprising over $1.6$ million schedules from $10,000$ pipelines. We focus on CPU-based hardware platforms and benchmark the schedules on Intel Xeon CPUs to obtain the run times for training.
    \item{\textbf{Improved Accuracy.}} An experimental evaluation of our approach shows the benefits of performance modeling using Graph Convolutional Networks (GCNs). On a random sampling of schedules, our model reduces the prediction error by a significant margin compared to the current Halide ($7.75x$) and TVM ($12x$) models. Moreover, our model is around $75\%$ accurate when performing a pair-wise ranking of schedules derived from real-world deep neural networks.
\end{itemize}

The rest of the paper is organized as follows. Section \ref{background} provides a background of the basic constructs of algorithm-schedule separation in Halide. We also provide a brief overview of the Halide~\cite{halide_auto} and TVM~\cite{tvm} performance models in this section. Section \ref{architecture} describes the proposed model architecture in detail. Section \ref{evaluation} presents the evaluation methodology and results. We discuss related work in Section \ref{relatedwork}, future work in Section \ref{future}, and conclude in Section \ref{conclusion}.

\section{Background}\label{background}

\subsection{Scheduling in Halide}
 Neural network computations are usually modeled in terms of operations involving tensors. This computational model gained popularity with the advent of deep learning libraries and engines like Tensorflow, PyTorch, etc. However, as discussed in the previous section, these frameworks rely on low-level optimized implementations of commonly used tensor operators when translating the computational graph of the network, and this may result in missed optimization opportunities across operators. The primary philosophy behind Halide is to separate the \emph{algorithm} (computation) from the \emph{schedule} (how to carry out the computation). For a deep-learning pipeline composed of one or more tensor operations, the Halide compiler can generate code for these operations such that it optimizes the overall throughput of the pipeline. Halide exposes programmers to a design space mostly hidden in traditional deep-learning frameworks.
 
 In this section, we provide a brief overview of the Halide framework and its scheduling primitives. As an example, we use Halide to implement a \emph{linear} layer. Linear layers are common in deep learning and represent a fully connected neuron layer.  The inputs to the layer are multiplied by a \emph{weight} matrix followed by the addition of \emph{bias} terms. In \emph{batched} computation, multiple input samples are processed concurrently. The following equation expresses the linear transformation of an input batch represented as a two dimensional matrix $X$, where each row represents an input. $W$ and $B$ are the weight and bias matrices, respectively.
 \begin{align*}
 Y = XW+B
 \end{align*}
 The following code snippet shows a Halide implementation of a linear layer. Please note that while this is not the best way to implement this operation, it is sufficient for our illustration. When doing \emph{batched} computations, the input and output matrices have the dimensions \emph{(batch\_size, input\_size)} and \emph{(batch\_size, output\_size)}, respectively. In this example, we use a batch of $64$, an input size of $1,024$, and an output comprising $16$ floats. \emph{ImageParam} is used to define a multi-dimensional input to the pipeline. The input, weight, and bias have two dimensions and store floating-point numbers.
 We split the computation across two functions or \emph{stages}, \emph{matrix\_mul} carries out the multiplication of the input with the weights, and \emph{add\_bias}, as the name suggests, sums the product with the \emph{bias} terms. The variables \emph{x} and \emph{y} are used as indices to specify the computation. The multiplication stage has two update definitions, the first one (\emph{line 15}) zeros the buffer, and the second update (\emph{line 16}) computes the product.
 The two matrices are multiplied using a \emph{reduction domain} (RDom) which denotes an implicit loop. For this two-stage pipeline, the \emph{add\_bias} stage consumes the result of the \emph{matrix\_mul} stage. The \emph{compute\_root} forces the completion of the \emph{matrix\_mul} stage before proceeding to the \emph{add\_bias} stage. Calling the \emph{realize} operation on the final stage compiles the entire pipeline.
\lstset{style=mystyle}
\begin{lstlisting}
const int batch = 64, input = 1024, output = 16;

ImageParam input(type_of<float>(),2); //batch*inputs
ImageParam wts(type_of<float>(),2);   //input*output
ImageParam bias(type_of<float>(),2);  //batch*output

Var x("x"), y("y");

Func matrix_mul("matrix_mul");
Func add_bias("add_bias");

RDom r(0, input);
RVar k = r[0];
matrix_mul(x, y) = 0.0f;
matrix_mul(x, y) += input(x, k) * wts(k, y);
add_bias(x, y) = matrix_mul(x, y) + bias(x, y);

Buffer<float> output(output, batch);
matrix_mul.compute_root();
add_bias.realize(output);
\end{lstlisting}
The rest of the section describes some typical scheduling options available in Halide. 
\subsubsection{Inline Evaluation}
This option, invoked using \emph{compute\_at}, inlines the computations performed in the \emph{producer} into the loop-nest of the \emph{consumer}. The producer function is computed as needed by the consumer function, which avoids the need for a temporary buffer to store the intermediate result. Inlining the evaluation also improves the \emph{locality of reference} as generated values are immediately consumed. A potential downside of inlining is that it may lead to redundant computations as intermediate results are not stored in memory.
The following code inlines the computations of the product stage into the inner loop of the bias stage.

\lstset{style=mystyle}
\begin{lstlisting}
matrix_mul.compute_at(add_bias, x);
\end{lstlisting}

\subsubsection{Reorder}
Loop reordering changes the hierarchy of the loop-nest. In certain situations, reordering can take advantage of data storage pattern to improve locality. The sequence in which the loop iteration variables provided to \emph{reorder} establishes the ordering beginning with the inner-most loop and ending at the outer-most loop. For example, the following line reorders the loop-nest of the \emph{add\_bias} stage. 
\lstset{style=mystyle}
\begin{lstlisting}
add_bias.reorder(y,x);
\end{lstlisting}
\subsubsection{Split}
The split functionality breaks a loop into a pair of nested loops. The \emph{split factor} is the trip count of the inner loop, and the outer loop extent is the original extent divided by the split factor. Using a combination of \emph{splitting} and \emph{reordering} can achieve \emph{tiled} execution or \emph{blocking}. The amount of cache memory is limited, and arrays seldom fit entirely in the cache. In such situations, \emph{tiling} partitions the iteration space into small chunks (working set) such that they fit in the cache, thus improving locality. The following code splits the x and y loops of the \emph{matrix\_mul} by a factor of 8. Reordering the outer four loops produces a partially tiled implementation.
\lstset{style=mystyle}
\begin{lstlisting}
Var x_outer, x_inner, y_outer, y_inner;
int split_factor = 8;
matrix_mul.update(0).split(x, x_outer, x_inner, split_factor);
matrix_mul.update(0).split(y, y_outer, y_inner, split_factor);
matrix_mul.update(0).reorder(x_inner, y_inner, x_outer, y_outer);
\end{lstlisting}
\subsubsection{Vectorize and Parallel}
The \emph{vectorize} option computes multiple iterations of a loop simultaneously by utilizing the available {SIMD} hardware. To parallelize a computation across CPU cores, we use the \emph{parallel} directive. For example, one way to parallelize and vectorize the \emph{add\_bias} stage would be to \emph{split} the inner loop by a factor of $4$ and \emph{vectorize} the \emph{x\_inner} loop and \emph{parallelize} the $y$ loop. The following code shows the relevant operations to implement this schedule.
\lstset{style=mystyle}
\begin{lstlisting}
add_bias.split(x, x_outer, x_inner, 4);
add_bias.vectorize(x_inner).parallel(y);
\end{lstlisting}

\subsubsection{Other options}
In addition to the options discussed in this section, Halide has several other scheduling knobs like loop-unrolling, loop fusions, storing intermediate results in a buffer, etc. We refer the reader to the Halide documentation~\cite{halide_doc} for further details. 

It should be evident from this discussion that Halide exposes the programmer to a rich and complex schedule space. For a single stage, the number of available scheduling options increases combinatorially with the number of tensor inputs, dimensions (\emph{rank}), and size of each dimension. 
As a result of this large search space, auto-tuning approaches that depend on benchmarking on actual hardware for feedback are simply too slow to be practical. This led both Halide and TVM to leverage performance models to assess the fitness of candidate solutions.

\begin{figure}
  \centering
  \begin{tikzpicture}
    \node (a) [state] {\small Deep Learning Application};
    \node (b) [state, right of=a, xshift=1.2cm] {\small Schedule Search};
    \node (c) [state, right of=b, xshift=1.2cm] {\small Candidate Schedules};
    \node (d) [state, right of=c, xshift=1.2cm] {\small Perf. Model};
    
    \draw [arrow] (a) -- (b);
    \draw [arrow] (b) -- (c);
    \draw [arrow] (c) -- (d);
    
    \draw [arrow] (d.north) -- ++(0,1) node[anchor=south, xshift=-2.2cm] {\small{Guides}} -| (b.north);
    
  \end{tikzpicture}
  \caption{\emph{Performance Model guided search:} For a given application, the model guides the search by selecting a subset of best performing schedules for further exploration.}
  \label{intro2}
\end{figure}
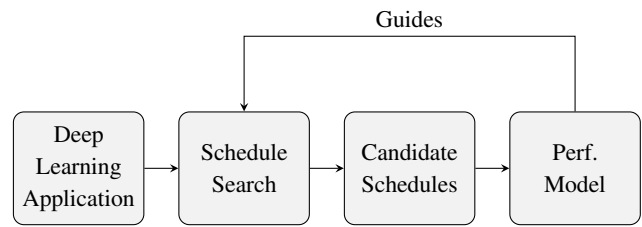
\subsection{Existing Performance Models}
\begin{figure}[h]
  \centering
  \begin{tikzpicture}
    \node[rectangle,draw,rounded corners, text width=2.5cm, text centered, fill=White] (1) {\small Algorithm \\ specific feat.};
    \node[rectangle,draw,rounded corners, right of=1, text width=2.5cm, text centered, fill=White, xshift=2.5cm] (2) {\small Schedule dependent feat.};
    \node[rectangle,draw,rounded corners, below of=1, text width=2.5cm, text centered, fill=White, yshift=-0.5cm] (3) {\small Embedding};
    \node[rectangle,draw,rounded corners, below of=2, text width=2.5cm, text centered, fill=White, yshift=-0.5cm] (4) {\small Embedding};
    \node[rectangle,draw,rounded corners, below of=3, text width=2.5cm, text centered, fill=White, xshift=1.75cm, yshift=-0.5cm] (5) {\small Stacked Embedding};
    \node[rectangle,draw,rounded corners, below of=5, text width=2.5cm, text centered, fill=White, yshift=-0.5cm] (6) {\small Coefficients};
    
    \draw [arrow] (1) -- (3);
    \draw [arrow] (2) -- (4);
    \draw [arrow] (3) |- (5);
    \draw [arrow] (4) |- (5);
    \draw [arrow] (5) -- (6);
        
  \end{tikzpicture}
  \caption{\emph{Halide auto-scheduler model: } The \emph{algorithm} and \emph{schedule} features are passed through fully connected layers to generate embeddings which are then combined and passed through a fully connected layer to get the final coefficients. The dot product of these coefficients with the 27 hand-crafted terms is the final run time~\cite{halide_auto}.}
  \label{halide_auto}
\end{figure}
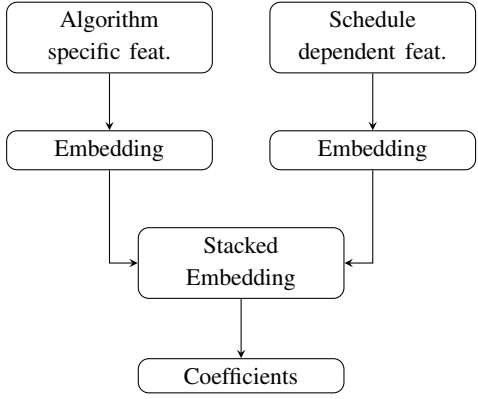
As  already  mentioned,  the  Halide  and  TVM  compilers
search the schedule space and find an efficient implementation of a given program. Figure \ref{intro2} depicts a basic search framework. The search technique generates a pool of candidate schedules and  uses  the  performance  model  to  select  the  most  promising  candidates  for  further  exploration.  
 In Halide, a computation comprises multiple \emph{stages} or \emph{functions}, and the search has to make a scheduling decision for each stage. In the Halide \emph{auto-scheduler}~\cite{halide_auto}, the stage-wise scheduling decisions are made using \emph{beam} search, which uses feedback from the performance model to maintain a pool of candidate schedules. To add an unscheduled stage, the search graph expands by enumerating all possible schedules for that stage. The model is then used to rank the resulting candidates and prune the graph preserving only the \emph{top-k} candidates. 

The model is constructed using fully connected layers and is trained to predict the run time conditioned on features extracted from the application. The featurization process results in two types of features. \emph{Algorithm-specific} features are invariant to the schedule and comprise a histogram of operations performed. These features characterize the computation and not the implementation. \emph{Schedule-dependent} features, on the other hand, capture the impact of the scheduling choices. These include metrics to capture the memory footprint, parallelization, vectorization, etc. The performance model computes low-dimensional embeddings of the algorithm-specific and schedule-dependent features and stacks them to generate the final output. 
Instead of directly predicting the run time, the output learns the coefficients of $27$ hand-crafted terms. These terms are derived from schedule-specific features. The network output assigns appropriate weights to each of the terms, and the run time is the dot product of the terms and coefficients.
Figure~\ref{halide_auto} is a high-level representation of the network used to construct the model. The model estimates the run time of each stage, and the summation of the run times of all scheduled stages provides the total run time or the performance of the pipeline~\cite{halide_auto}.

The TVM auto-scheduler employs a \emph{gradient boosted tree} (GBT) model based on \emph{XGBoost}~\cite{tvm_model}. The inputs to the model are features that encode information about the loop nest. These context features include information about the loop structure like the loop extents, memory footprint, and annotations for vectorization, unrolling, etc. To make the context features more generalizable, the authors build \emph{context relation} features which model the relationship between different features in the loop nest that affect the performance. TVM also has a second model built using \emph{TreeGRUs}. Instead of directly learning an embedding for each identifier in the loop nest AST, the model uses embeddings derived from the GBT \emph{context vectors}. In practice, both models exhibit similar performance. However, the GBT model is faster than the TreeGRU model~\cite{tvm}.

\subsection{Feature Engineering}
The proposed work follows a methodology similar to the previous deep-learning based Halide models and extracts two categories of features from Halide programs~(see \cite{halide_auto, value_learn}). The first category of features, called \emph{pipeline} or \emph{schedule invariant} features, intends to characterize the nature of the computation. The second category, called \emph{schedule dependent} features, captures how the computations are performed.
\subsubsection{Schedule-invariant Features}
Schedule-invariant features describe the computations and are independent of how they are carried out. As the name suggests, these features remain consistent across different schedules of the same pipeline. We compute a histogram of different types of operations performed to produce the result. These include floating-point arithmetic operations on tensors as well the integer arithmetic used for tensor indexing. We also capture boolean/logical operations like and, or, xor, etc. 
Memory access latencies have a significant impact on execution times. We also record access patterns like striding behavior, transposed access, and broadcasts. 
\subsubsection{Schedule-dependent Features}
This set of features capture the scheduling choices made to execute a pipeline. Modern CPUs exhibit considerable variation in resources like the number of ALU units, cache sizes and hierarchy, memory bandwidth, number of cores, etc. An optimal schedule organizes the computation such that all available resources are utilized to the maximum extent possible. The pipeline is scheduled stage-by-stage, beginning from the last/output stage and going up the DAG to the source/input stage(s). Information about scheduling decisions for every stage is encoded using the following data points. Loop-splitting transforms a loop into a nested configuration with one outer and one inner loop. This scheduling option has implications on the locality of reference and determines the efficiency of cache utilization. The impact of splitting a loop is captured by recording the new loop ranges/extents. The memory footprint of each loop is quantified by counting the number of unique cache lines accessed, histogram of accessed bytes, and memory reuse distance. A lot of current CPUs exploit data-level parallelism via SIMD/vector instructions. We count the number of vectorized and scalar floating-point and integer arithmetic operations. The CPU core utilization is measured as a ratio of parallel tasks and the total cores. The impact of inlining a function call is measured in terms of the additional computation performed. Schedule-dependent features also incorporate an estimate of other overheads that impact performance. These include the amount of allocations and deallocations on the heap, context switches, and page faults.
\newline
\textbf{Compound Features: } 
The Halide-based model introduced in \cite{value_learn} used a subset of the schedule-dependent features to capture products and ratios, which are difficult for a network to learn without significantly bloating network capacity. The authors augmented the schedule-dependent features with these compound features derived from two or more basic features via comprehensive evaluation. These include features that quantify the impact of memory allocations, page faults, arithmetic intensity, to name a few. We also incorporate these features in our model.

\section{Architecture}\label{architecture}

This section describes the design and implementation of our model. To develop our model, we follow the standard \emph{supervised} learning approach and use a corpus of randomly generated Halide pipelines and schedules for training. We begin this section by detailing the dataset generation methodology. We then explain the construction of the performance model and its building blocks.

\subsection{Dataset Generation}
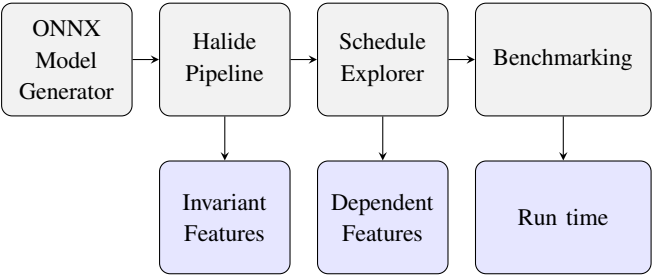
\begin{figure}
	\begin{tikzpicture}
    \node (a) [state] {\small ONNX Model Generator};
	\node (b) [state, right of=a, xshift=1.1cm] {\small Halide Pipeline};
    \node (c) [state, right of=b, xshift=1.1cm] {\small Schedule Explorer};
    \node (d) [state1, right of=c, xshift=1.4cm] {\small Benchmarking};
    \node (e) [state, below of=b, yshift=-1.1cm,  fill=blue!10] {\small Invariant Features};
    \node (f) [state, below of=c, yshift=-1.1cm,  fill=blue!10] {\small Dependent Features};
    \node (g) [state1, below of=d, yshift=-1.1cm, fill=blue!10] {\small Run time};
    \draw [arrow] (a) -- (b);
    \draw [arrow] (b) -- (c);
    \draw [arrow] (c) -- (d);
    \draw [arrow] (b) -- (e);
    \draw [arrow] (c) -- (f);
    \draw [arrow] (d) -- (g);
	\end{tikzpicture}
 	\caption{\emph{Data generation pipeline:} Randomly generated ONNX models are converted to Halide pipelines, and the schedule-independent features are extracted at this stage. The pipelines are scheduled using the Halide auto-scheduler, and the schedules are benchmarked on the target hardware to record the actual run times.}
  	\label{datagen}
\end{figure}

The data generation framework depicted in figure 4 has four components. First, we use a random model generator to build models conforming to the ONNX~\cite{onnx} standard. The ONNX models are then converted to Halide pipelines. Next, we obtain multiple schedules for each randomly generated model, and finally, we benchmark the schedules to get the run time. The rest of this section describes the components in detail. 

Algorithm \ref{algo} shows the data generation process.
The random model generator constructs models by using operators commonly found in deep learning. These include operations like Gemm, Conv, Maxpool, Average Pool, Relu, Sigmoid, Softmax, etc. We have identified about $50$ such operators. We now describe the procedure to generate random ONNX models. ONNX, short for Open Neural Network Exchange, is an open standard to specify neural network models that can be exported across a wide variety of environments and platforms~\cite{onnx}. After choosing the number of inputs to the model (\emph{line 3}), we generate the input tensors (\emph{line 4}). The tensor ranks and dimensions are sampled from a specified range. The set of inputs comprise the input stage of the model. The next step is to determine the number of layers in the model (\emph{line 5}). 
We build each network stage by stage, with the outputs of the previous stage acting as inputs to the current one. (\emph{lines 8-10}). Each stage comprises several computational nodes, and their number is randomly sampled from a range (\emph{line 23}). We support different kinds of nodes depending on the number of inputs (\emph{node.type}) and type of operation (\emph{node.op}). When building a random node, we sample from predefined distributions to determine the type and the operation (\emph{lines 31, 35, 38}). After building all the nodes, we copy the unused tensors (tensors that weren't compatible inputs to any node) from the input stage to the next stage (\emph{line 27}). To ensure the collection of only realistic networks, we add two filters that discard most graphs with more than one output (\emph{$output\_thresh=1$}) and graphs with a depth of less than $5$ (\emph{$depth\_thresh=5$}). A large number of well-known deep neural networks make use of operators like convolutions, Relu activations, etc. (\emph{favored\_ops}). 
To obtain a more representative distribution of models, we filter a large number of networks lacking such operators (\emph{lines 15-16}). We return the model only if it passes all the filters (\emph{lines 17-20}). The random ONNX pipelines are then translated into Halide pipelines. At this stage, we extract the schedule-invariant features. 

The next step is to generate schedules for the pipelines for which we utilize the Haldide auto-scheduler~\cite{halide_auto}. By injecting the performance model with random noise, we can derive multiple schedules for each pipeline. For every schedule, we record the schedule-dependent features. Finally, for benchmarking, each schedule is run $10$ times on the target hardware to obtain the run time. Our dataset comprises over $1.6$ million schedules from around $10,000$ pipelines. We use $10\%$ of the dataset for evaluation and the rest for training. The schedules were benchmarked on 18 core Intel Xeon CPUs (D-2191) running at 1.60 GHz and having a memory of 48GB.

\begin{algorithm}
  \caption{ONNX Model Generator}
  \begin{algorithmic}[1]
  \Function{build\_random\_ONNX\_model}{}
    \State $model \leftarrow \text{initialize model}$
    \State $num\_inputs \leftarrow \text{determine no. of inputs}$
    \State $input\_stage \leftarrow generate\_inputs(num\_inputs, ...)$
    \State $num\_stages \leftarrow \text{number of stages in the pipeline}$
    \newline\newline \Comment{Add stages one by one}
    \While{$\text{not done generating } num\_stages$} 
        \State $new\_stage \leftarrow build\_new\_stage(input\_stage, ...)$
        \State $\text{add } new\_stage \text{ to } model $
        \State $input\_stage \leftarrow new\_stage$
    \EndWhile
    \newline
    \State $valid\_outputs \leftarrow $
    \State $\hspace*{1.5em}filter\_outputs(model.outputs, output\_thresh)$
    \newline
    \State $valid\_depth \leftarrow $
    \State $\hspace*{1.5em}filter\_depth(model, depth\_thresh)$
    \newline
    \State $favored\_ops = \text{\{conv, relu, ...\}}$
    \State $model\_OK \leftarrow $
    \State $\hspace*{1.5em}filter\_model(model, favored\_ops)$
    \newline
    \If{$valid\_outputs$ and $valid\_depth$ \newline \hspace*{1.5em} and $model\_OK$}
        \State $\text{\textbf{return }}model$
    \Else 
        \State $\text{\textbf{return }}invalid\_model$
    \EndIf
  \EndFunction
  \newline
  \Function{build\_new\_stage}{input\_stage, ...}
    \State $new\_stage \leftarrow \text{ list containing nodes in this stage}$
    \State $width \leftarrow \text{ number of nodes in this stage}$
    \newline\newline \Comment{Add nodes to the stage}
    \While{$\text{not done generating } width \text{ nodes}$} 
        \State $new\_node \leftarrow build\_random\_node(input\_stage, ...)$
        \State $\text{add } new\_node \text{ to } new\_stage $
    \EndWhile
    \newline
    \State $\text{add unused tensors from } input\_stage \text{ to } new\_stage$     
    \State $\text{\textbf{return }}new\_stage$ 
  \EndFunction
  \newline
  \Function{build\_random\_node}{input\_stage, ..}
  \State $node \leftarrow \text{initialize node}$
  \State $node.type \leftarrow $
  \State $\hspace*{5em}sample\_categorical(unary, binary, ...)$
  \newline
  \If{$node.type$ is $unary$}
    \State $node.op \leftarrow $ 
    \State $\hspace*{2em}sample\_categorical(pad, pool, softmax, ...) $
  \ElsIf{$node.type$ is $binary$}
    \State $node.op \leftarrow $ 
    \State $\hspace*{0.5em}sample\_categorical(conv, gemm, batch\_norm, ...) $
  \EndIf
  \State $...$
  \State $\textbf{return } node$
  \EndFunction
  \end{algorithmic}
  \label{algo}
\end{algorithm}

\subsection{Model}
\begin{figure}
  \centering
  \begin{tikzpicture}
    \node[rectangle,draw,rounded corners, text width=2.5cm, text centered, fill=White] (stage) {\small Embedding (280)};
    \node[trapezium,draw,rounded corners, below of=stage, text width=2cm, xshift=-1.75cm, yshift=-0.2cm, text centered, fill=White] (lin1) {\small Linear (220)};
    \node[trapezium,draw,rounded corners, below of=stage, text width=2cm, xshift=1.75cm, yshift=-0.2cm, text centered, fill=White] (lin2) {\small Linear (60)};
    \node[trapezium,draw,rounded corners, below of=lin1, text width=2cm, xshift=0cm, yshift=-0.2cm, text centered, fill=White] (norm1) {\small Norm. };
    \node[trapezium,draw,rounded corners, below of=lin2, text width=2cm, xshift=0cm, yshift=-0.2cm, text centered, fill=White] (norm2) {\small Norm. };
    \node[rectangle,draw,rounded corners, below of=norm1, text width=2.7cm, xshift=0cm, yshift=-0.2cm, text centered, fill=White] (dep) {\small Invariant (320)};
    \node[rectangle,draw,rounded corners, below of=norm2, text width=2.7cm, xshift=0cm, yshift=-0.2cm, text centered, fill=White] (indep) {\small Dependent (94)};
    \node[rectangle,draw,rounded corners, below of=stage, yshift=-3.8cm, text width=2.5cm, text centered, fill=White] (emb) {\small Stage (Node)};
    
    \draw [arrow] (indep) -- (norm2);
    \draw [arrow] (dep) -- (norm1);
    \draw [arrow] (norm1) -- (lin1);
    \draw [arrow] (norm2) -- (lin2);
    \draw [arrow] (lin1) -- (stage);
    \draw [arrow] (lin2) -- (stage);
    \draw [arrow] (emb) -- (indep);
    \draw [arrow] (emb) -- (dep);
    
  \end{tikzpicture}
  \caption{\emph{Feature Generation and Embedding:} The schedule invariant and dependent features are extracted from every stage, normalized, and then passed through linear layers to generate the respective embeddings. The embeddings are combined to derive the feature vector for the stage. The number in the brackets denotes the size of the feature/embedding.}
  \label{feat}
\end{figure}
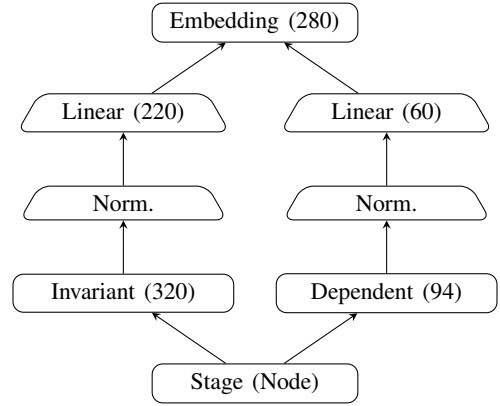
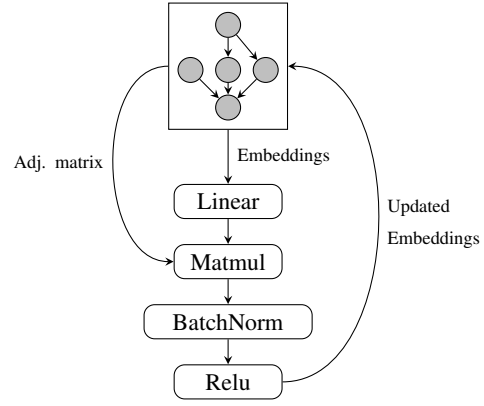
\begin{figure}
  \centering
  \begin{tikzpicture}
    \node[circle,draw,fill=gray!50, xshift=0.9cm] (c1) {};
    \node[circle,draw,fill=gray!50, below of=c1, yshift=0.4cm] (c2) {};
    \node[circle,draw,fill=gray!50, right of=c2, xshift=-0.5cm] (c3) {};
    \node[circle,draw,fill=gray!50, left of=c2, xshift=0.5cm] (c4) {};
    \node[circle,draw,fill=gray!50, below of=c2, yshift=0.5cm] (c5) {};
    \node (conv_box) [draw,fit= (c1) (c2) (c3) (c4) (c5)] {};
    
    \node (linear2) [stage, below of=conv_box, draw=black, yshift=-0.8cm, xshift=-0.0cm] {\small Linear};
    \node (matmul) [stage, below of=linear2, draw=black, yshift=0.2cm, xshift=0.0cm] {\small Matmul};
    \node (bn) [stage, below of=matmul, draw=black, text width=2cm, yshift=0.2cm] {\small BatchNorm};
    \node (relu) [stage, below of=bn, draw=black, yshift=0.2cm] {\small Relu};
    
    \draw [arrow] (linear2) -- (matmul);
    \draw [arrow] (matmul) -- (bn);
    \draw [arrow] (bn) -- (relu);
    
    \draw [arrow] (conv_box.west) to [out=180, in=180] node[anchor=east, text width=2cm, xshift=0.8cm] {\scriptsize{Adj. matrix}} (matmul.west);
    \draw [arrow] (conv_box) -- node[anchor=west] {\scriptsize{Embeddings}} (linear2);
    \draw [arrow] (c1) -- (c2);
    \draw [arrow] (c1) -- (c3);
    \draw [arrow] (c2) -- (c5);
    \draw [arrow] (c3) -- (c5);
    \draw [arrow] (c4) -- (c5);
    \draw [arrow] (relu.east) to [out=0, in=0] node[anchor=west, text width=2cm] {\scriptsize{Updated Embeddings}} (conv_box.east);
    
  \end{tikzpicture}
  \caption{\emph{Graph Convolutional Block:} The linear layer takes as input the current embeddings of the graph nodes, and the output is multiplied with the normalized adjacency matrix self-loop edges. The product is then batch-normalized and passed through \emph{Relu} activation, which generates the new node embeddings.}
  \label{conv}
\end{figure}
Graphs are an effective way to model a collection of objects and their relationships. Machine learning formulations for graph-structured data have found application and success in the several domains~\cite{gnn_review}. Deep learning networks that process data modeled as graphs have been applied to learn representation for nodes, the relationship between nodes (edges), clusters (subgraphs), and entire graphs. Our approach models the performance prediction problem as a regression task on learned graph representations. There is an extensive body of work in graph-based learning, and we determined the \emph{Graph Convolutional Network}~\cite{gcn} framework to be the most suitable for our specific applications scenario. 
\begin{figure*}
	\centering
	\definecolor{c1}{HTML}{f54295}
	\definecolor{c2}{HTML}{f5f542}
	\definecolor{c3}{HTML}{66FF66}
    \definecolor{c4}{HTML}{00CCFF}
    
    \definecolor{c5}{HTML}{c5bd6a}
    \definecolor{c6}{HTML}{33e6b3}
    
    \definecolor{c7}{HTML}{7cd28f}
    
    \definecolor{c8}{HTML}{94c18f}
    \definecolor{c9}{HTML}{b9b77d}
    \definecolor{c10}{HTML}{cbb274}
    
	\begin{tikzpicture}
    \node (a) [stage, fill=c1!80] {Stage 1};
	\node (b) [stage, right of=a, xshift=0.9cm, fill=c2!80] {Stage 2};
    \node (c) [stage, below of=b, xshift = -0.9cm, yshift=-0.2cm, fill=c3!80] {Stage 3};
    \node (d) [stage, below of=c, yshift=-0.2cm, fill=c4!80] {Stage 4};
    \draw [arrow] (a) -- (c);
    \draw [arrow] (b) -- (c);
    \draw [arrow] (c) -- (d);
    
    \node (e) [stage, fill=c1!80, right of=b, xshift=2.5cm] {Stage 1};
	\node (f) [stage, right of=e, xshift=0.9cm, fill=c2!80] {Stage 2};
    \node (g) [stage, below of=f, xshift = -0.9cm, yshift=-0.2cm, fill=c5!80] {Stage 3};
    \node (h) [stage, below of=g, yshift=-0.2cm, fill=c6!80] {Stage 4};
    \draw [arrow] (e) -- (g);
    \draw [arrow] (f) -- (g);
    \draw [arrow] (g) -- (h);
    
    \node (i) [stage, fill=c1!80, right of=f, xshift=2.5cm] {Stage 1};
	\node (j) [stage, right of=i, xshift=0.9cm, fill=c2!80] {Stage 2};
    \node (k) [stage, below of=j, xshift = -0.9cm, yshift=-0.2cm, fill=c5!80] {Stage 3};
    \node (l) [stage, below of=k, yshift=-0.2cm, fill=c7!80] {Stage 4};
    \draw [arrow] (i) -- (k);
    \draw [arrow] (j) -- (k);
    \draw [arrow] (k) -- (l);
    
    \node (c1) [conv, right of=c, yshift=2.7cm, xshift=-1cm] {convolution};
    \node (c2) [conv, right of=g, yshift=2.7cm, xshift=-1cm] {convolution};
    \node [single arrow, draw=black, minimum height=2em, right of=c, xshift=0.5cm, fill=black] {};
    \node [single arrow, draw=black, minimum height=2em, right of=c1, xshift=0.2cm, fill=black] {};
    \node [single arrow, draw=black, minimum height=2em, right of=g, xshift=0.5cm, fill=black] {};
    \node [single arrow, draw=black, minimum height=2em, right of=c2, xshift=0.2cm, fill=black] {};
    
    \node (con2_1) [stage2, minimum height = 0.4cm, fill=c1!80, below of=h, yshift=-1.0cm, xshift=-5cm] {};
	\node (con2_2) [stage2, minimum height = 0.4cm, below of=con2_1, fill=c2!80, yshift=0.5cm] {};
    \node (con2_3) [stage2, minimum height = 0.4cm, below of=con2_2, fill=c5!80, yshift=0.5cm] {};
    \node (con2_4) [stage2, minimum height = 0.4cm, below of=con2_3, fill=c6!80, yshift=0.5cm] {};
    
    \node (con1_1) [stage2, minimum height = 0.4cm, fill=c1!80, left of=con2_1, xshift=-0.3cm] {};
	\node (con1_2) [stage2, minimum height = 0.4cm, below of=con1_1, fill=c2!80, yshift=0.5cm] {};
    \node (con1_3) [stage2, minimum height = 0.4cm, below of=con1_2, fill=c3!80, yshift=0.5cm] {};
    \node (con1_4) [stage2, minimum height = 0.4cm, below of=con1_3, fill=c4!80, yshift=0.5cm] {};
    
    \node (con3_1) [stage2, minimum height = 0.4cm, fill=c1!80, right of=con2_1, xshift=0.3cm] {};
	\node (con3_2) [stage2, minimum height = 0.4cm, below of=con3_1, fill=c2!80, yshift=0.5cm] {};
    \node (con3_3) [stage2, minimum height = 0.4cm, below of=con3_2, fill=c5!80, yshift=0.5cm] {};
    \node (con3_4) [stage2, minimum height = 0.4cm, below of=con3_3, fill=c7!80, yshift=0.5cm] {};
    
    \node [draw, below of=d, yshift=0.2cm, draw=white] {Initial Embeddings ($E^0$)};
    \node [draw, below of=h, yshift=0.2cm, draw=white] {After 1st Conv. ($E^1$)};
    \node [draw, below of=l, yshift=0.2cm, draw=white] {After 2nd Conv. ($E^2$)};
    
    \node [draw, below of=con2_4, yshift=0.3cm, draw=white] {$E^0$, $E^1$, and $E^2$};

    
    
    \node [single arrow, draw=black, minimum height=2em, right of=con3_2, xshift=0.2cm, yshift=-0.25cm, fill=black] {};
    
    \node (sum_block) [conv, right of=con3_2, yshift=2.2cm, xshift=-0.72cm] {sum pool};
    
    \node (b_arr5) [single arrow, draw=black, minimum height=2em, right of=sum_block, xshift=-0.1cm, yshift=-0.00cm, fill=black] {};
    
    \node (sum1) [stage3, right of=b_arr5, fill=c8!80, yshift=0.0cm, xshift=0.0cm] {};
    \node (sum2) [stage3, above of=sum1, fill=c9!80, yshift=-0.1cm, xshift=0.0cm] {};
    \node (sum3) [stage3, below of=sum1, fill=c10!80, yshift=0.1cm, xshift=0.0cm] {};
    
    \node [draw, below of=sum3, yshift=0.0cm, draw=white, text width=2.4cm] {Consolidated Embeddings ($F$)
    };
    
    \node (b_arr6) [single arrow, draw=black, minimum height=2em, right of=sum1, xshift=-0.1cm, yshift=-0.00cm, fill=black] {};
    
    \node (linear1) [conv, right of=b_arr6, yshift=1.0cm, xshift=-1.0cm] {Linear ($W_{out}$)};
    
    \node (b_arr7) [single arrow, draw=black, minimum height=2em, right of=linear1, xshift=-0.1cm, yshift=-0.00cm, fill=black] {};
    
    \node [draw, right of=b_arr7, xshift=0.5cm, draw=white] {\emph{run time ($\hat{y}$)}};
    
    \end{tikzpicture}
 	\caption{\emph{Model Architecture:} The initial node embeddings are passed through two convolutional layers. The initial ($E^0$), intermediate ($E^0$), and final embeddings ($E^0$), are preserved, and their stage-wise sum ($F$) is passed through a linear layer to produce the run time ($\hat{y}$).}
  	\label{gcn}
\end{figure*}
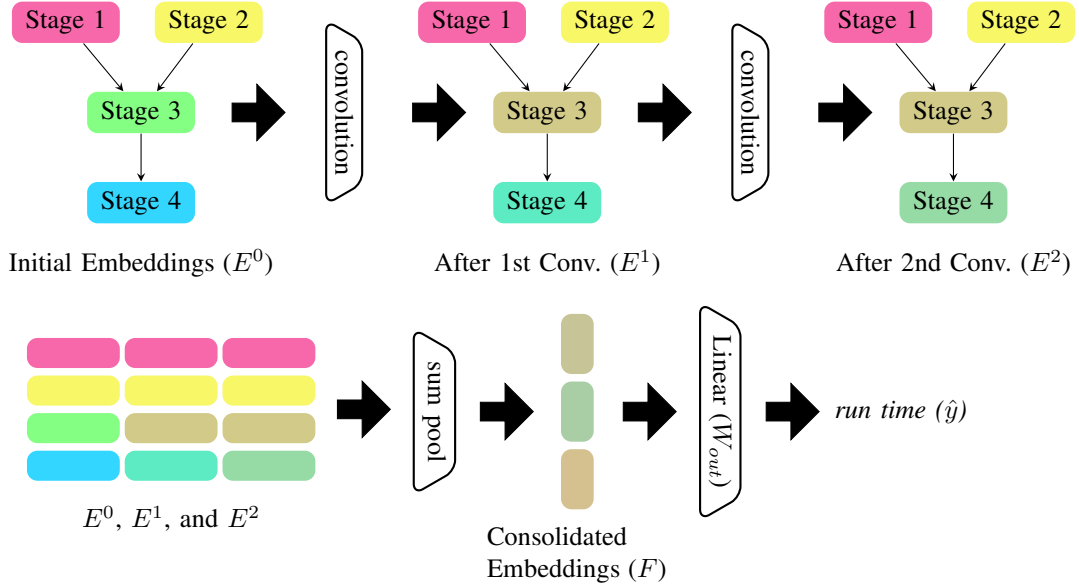
We now provide an overview of the our graph-based model. The following equation describes a directed graph where $V$ denotes the set of vertices and $E$ is the set of edges. 
\[G = (E,V) \] 
Each vertex $i$ in $V$ is represented by a feature vector $e$. In our case, each node is a Halide stage, and the initial embeddings ($e^0_i$) are derived as shown in figure \ref{feat}. To  generate  the  initial  embeddings, we  normalize the schedule-invariant and dependent features over the entire training set and embed the features to a low dimensional space for more efficient training.
\[e_i^0 = f_{init}(\text{\small{stage}}_{sched\_inv}, \text{\small{stage}}_{sched\_dep})\]
The graph connectivity and structure are encoded by the adjacency matrix $A$. We do not differentiate between edge types and do not have a feature vector for the edges. The network is built using two standard computational modules.
\newline\newline
\textbf{Aggregation-Update: }
This operation involves the nodes propagating their information to the neighboring nodes.
Each node aggregates the information it receives from its neighborhood and uses it to update its embeddings.
Graph networks typically have multiple aggregation-update steps. At the $k^{th}$ round of aggregation-update, to compute the next set of embeddings for some vertex $i$, the function $f_{aggregate}$ takes as input the current embeddings of $i's$ neighbors ($N(i,k)$) and consolidates them to generate the aggregated result $m_i^k$.
\[m_i^k = f_{aggregate}(N(i, k))\]
The function $f_{update}$, using this result and current embeddings of node $i$, generates the new embedding.
\[e_i^{k+1} = f_{update}(e_i^k, m_i^k)\]
The \emph{aggregate} operation is implemented by summing the embeddings of the neighboring nodes and passing the sum through a \emph{learnable weight matrix}. For the \emph{update}, we multiply the current embeddings of the node with another learnable matrix and add it to the output from the \emph{aggregate} operation. The final result is passed through an element-wise non-linearity. Along with the weight matrix, we also incorporate learnable biases but exclude them here for simplicity. The reasoning behind the update operation is that the new node representations will depend on the representations of the neighboring nodes as well as their own current states. The equation below shows the combined \emph{aggregate-update} operation.
\[e_i^{k+1} = \sigma(W_{self}^k.e_i^k + W_{agg}^k.\sum N(i,k))\]
We can transform the above equation, which captures these operations for individual nodes, to one representing the same operations on the graph level. $E$ denotes the embedding matrix of all nodes at step $k$, with each row corresponding to a graph node. Multiplication of the adjacency matrix with $E$ results in the summing of the neighbors for every node, and multiplying the result with the weight matrix achieves the aggregation operation. Similarly, we can multiply $E$ with $W_{self}^k$ to compute the updates for all nodes.
\[E^{k+1} = \sigma(W_{self}^k.E^k + A.E^k.W_{agg}^k)\]
Finally, we simplify the above equation along the lines of the \emph{graph convolutions} proposed by Kipf and Welling~\cite{gcn}. In their technique, the two different weight matrices are replaced by a single learnable matrix. The adjacency matrix is modified to incorporate self-loops by adding it to an \emph{identity} matrix. The resulting matrix is then row-normalized (each row sums to $1$), and multiplication with the transformed adjacency matrix ($A^{'}$) results in averaging the representation of the neighborhood nodes. Each aggregation-update step is referred to as a \emph{convolution} layer, similar in spirit to traditional \emph{convolutional networks}, with all nodes sharing the same set of weights. 
\[E^{k+1} = \sigma(A^{'}.E^k.W^k)\]
\textbf{Pooling-Readout:}
 After passing the initial node representations through one or more convolutional layers, we use the initial, intermediate, and final embeddings to learn representations for the entire graph. The pooling operation involves summing the node representations. Following a technique similar to~\cite{DGCNN}, we preserve the initial, intermediate node representations along with the final representations. The \emph{sum-pool} operation involves summing the node representations at every convolution step, including the initial embeddings. 
In the following equation, $F(k)$ is the sum of all node embeddings after the $k^{th}$ convolution.
\[F(k) = \sum\limits_{i\in V} e^k_{i}\]
The pooling operation produces a consolidated feature vector $F$ whose width is equal to the width of node embeddings multiplied by the number of convolutional layers. To generate the final readout or output ($\hat{y}$), the \emph{runtime} in our case, we pass the feature vector through a learnable weight matrix ($W_{out})$.
\[\hat{y} = W_{out}.F\]

\subsection{Implementation}
The initial node representations are generated by normalizing the schedule invariant and dependent features and embedding them to a lower-dimensional space (figure \ref{feat}).  Figure \ref{conv} shows a single convolution layer. The current embeddings of every node are passed through a linear layer, and multiplication of the output with a normalized adjacency matrix with self-loop achieves neighborhood \emph{aggregation} and \emph{update}. The addition of a \emph{batch-normalization} layer followed by a \emph{Relu} activation to generate the new embeddings results in better performance in practice. Our performance model, as seen in figure \ref{gcn}, comprises two convolutional layers. We arrived at this configuration after a parametric sweep of convolutional layers ranging from 0 to 8. Using the technique proposed in \cite{DGCNN}, we preserve the initial and intermediate embeddings for graph-level readout. To generate a representation for the entire graph, we sum across stages after every convolution and use a linear layer to obtain the runtime. We used the Adagrad~\cite{adagrad} optimizer with a \emph{learning rate} of \emph{0.0075} and a \emph{weight decay} of \emph{0.0001}.

\textbf{Loss Function:} We made the following choices to  design our loss function. 
\begin{itemize}
\item{\textbf{Property 1. }} We base our loss on the  the absolute ratio of the predicted and the measured run times. To account for noise in the measurements, every schedule $s$ of a pipeline $p$ is benchmarked $N$ times on the hardware, and the term $y_{ps}$ represents the mean of the measurements. N is set to 10 for the experiments in this paper. 
The following equation then determines the absolute relative error, where $\hat{y}_{ps}$ is the output of the model.
\[\xi = \abs{ \dfrac{N.\hat{y}_{ps}}{\sum\limits_{i=1}^{N} y_\text{$ps\_i$}} } \]
\item{\textbf{Property 2. }} We also want the model to differentiate between good and bad schedules. Making an accurate prediction on an efficient schedule is more important than making an accurate prediction on an inefficient schedule. To capture this behavior, we introduce a term $\alpha$ which, for a given pipeline $p$ and schedule $s$, is the ratio of the run time of the best schedule for $p$ and the run time for the current schedule $s$.
\[\alpha = \dfrac{\text{\small{$min(Schedules(p))$}}}{y_{ps}} \]
\item{\textbf{Property 3. }} The loss function should give more importance to less noisy measurements compared to measurements that have more noise. To embody this property, we define $\beta$ as the inverse of the standard deviation of measurements of a schedule $s$ of a pipeline $p$.
\[\beta = \dfrac{1}{std\_dev(y_\text{ps\_1}, y_\text{ps\_2}, ..., y_\text{ps\_N})} \]
\end{itemize}
We combine the three terms to compute the final loss, as shown in the equation below.
\[\ell_{ps} = \xi.\alpha.\beta \]

\section{Evaluation}\label{evaluation}

This section explains the evaluation methodology and presents our findings and results. We assess the prediction accuracy of the proposed performance model on the test dataset and compare it with the Halide and TVM models. We also evaluate the model's performance on ranking schedules derived from nine well-known deep-learning networks.

\subsection{Prediction Accuracy}
\begin{figure}[h]
\centering
    \begin{subfigure}[h]{\columnwidth}
        \begin{tikzpicture}
            \begin{axis}[ 
                xbar, 
                xmin=0,
                axis x line=none,
                axis y line=left,
                enlarge y limits=0.25,
                xlabel={Percentage \%},
                symbolic y coords={%
                    {TVM},
                    {Halide},
                    {Our}},
                ytick=data,
                height=3.5cm,
                width=\columnwidth,
                nodes near coords, 
                nodes near coords style={color=black},
                ytick=data,
            ]
            \addplot[color=OliveGreen!60, fill=OliveGreen!60] coordinates {
                (8.2,{Our})
                (64,{Halide})
                (104,{TVM})};
            \end{axis}
        \end{tikzpicture}
        \caption{\emph{Average Error(\%)} (lower is better) Percentage error between the measured (actual) and predicted run times computed for the test set.}
        \label{avg_error}
    \end{subfigure}
    \par\bigskip
    \begin{subfigure}[h]{\columnwidth}
        \begin{tikzpicture}
            \begin{axis}[ 
                xbar, 
                xmin=0,
                axis x line=none,
                axis y line=left,
                enlarge y limits=0.25,
                xlabel={Percentage \%},
                symbolic y coords={%
                    {TVM},
                    {Halide},
                    {Our}},
                ytick=data,
                height=3.5cm,
                width=\columnwidth,
                nodes near coords, 
                nodes near coords style={color=black},
                ytick=data,
            ]
            \addplot[color=TealBlue!60, fill=TealBlue!60] coordinates {
                (278,{Our})
                (4473,{Halide})
                (55109,{TVM})};
            \end{axis}
        \end{tikzpicture}
        \caption{\emph{Maximum Error(\%)} (lower is better) The maximum percentage error recorded.}
        \label{max_error}
    \end{subfigure}
    \par\bigskip
    \begin{subfigure}[h]{\columnwidth}
        \begin{tikzpicture}
            \begin{axis}[ 
                xbar, 
                xmin=0,
                axis x line=none,
                axis y line=left,
                enlarge y limits=0.25,
                xlabel={Percentage \%},
                symbolic y coords={%
                    {TVM},
                    {Halide},
                    {Our}},
                ytick=data,
                height=3.5cm,
                width=\columnwidth,
                nodes near coords, 
                nodes near coords style={color=black},
                ytick=data,
            ]
            \addplot[color=BrickRed!60, fill=BrickRed!60] coordinates {
                (0.92,{Our})
                (0.89,{Halide})
                (0.81,{TVM})};
            \end{axis}
        \end{tikzpicture}
        \caption{\emph{$R^{2}$} (higher is better) \emph{$R^{2}$} score or the \emph{coefficient of determination} quantifies how well the model fits the observational data. With values between 0 and 1, a higher \emph{$R^{2}$} score indicates a better fit.}
        \label{r_sq}
    \end{subfigure}
    \caption{Quality of \emph{our} performance model vs. \emph{Halide} and \emph{TVM} models}
    \label{error_plots}
\end{figure}
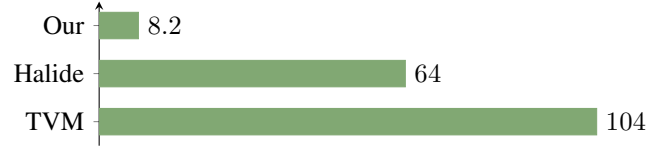
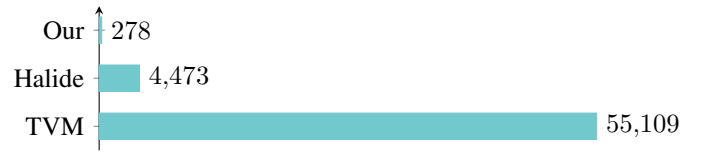
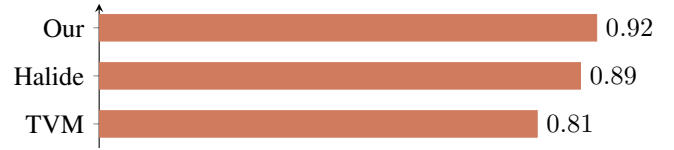
\pgfplotstableread[row sep=\\,col sep=&]{
    interval    & ranking \\
    alexnet     & 80  \\
    mnasnet     & 65 \\
    resnet18    & 80 \\
    resnet50    & 80 \\
    shufflenet  & 65  \\
    squeezenet  & 80 \\
    transformer & 65 \\
    vgg19       & 75 \\
    wavenet     & 90 \\
    }\mydata
    
\begin{figure*}[ht]
    \centering
    \begin{tikzpicture}
            \begin{axis}[
                ybar,
                symbolic x coords={alexnet, mnasnet, resnet18, resnet50, shufflenet, squeezenet, transformer, vgg19, wavenet},
                xtick=data,
                width=17.5cm,
                bar width=0.7cm,
                height=6cm,   
                ymin=0,
                ymax=100,
                grid=major,
                ylabel={Pairs correctly ranked($\%$)},
                ytick = {0, 10, 20, 30, 40, 50, 60, 70, 80, 90, 100},
                set layers,
                cell picture=true,
                extra y tick labels={},
                extra y tick style={
                    ymajorgrids=true,
                    ytick style={
                        /pgfplots/major tick length=0pt,
                    },
                    grid style={
                        black,
                        dashed,
                        /pgfplots/on layer=axis foreground,
                    },
                },
            ]
            \addplot table[x=interval,y=ranking]{\mydata};
        \end{axis}
    \end{tikzpicture}
    \caption{Ranking performance of the proposed model (higher is better): For all possible pairwise combinations of schedules belonging to a particular network, we count the number of pairs in which our model correctly identified the faster schedule. For each network, we plot the percentage of correctly ranked pairs along the y-axis.}
    \label{ranking}
\end{figure*}
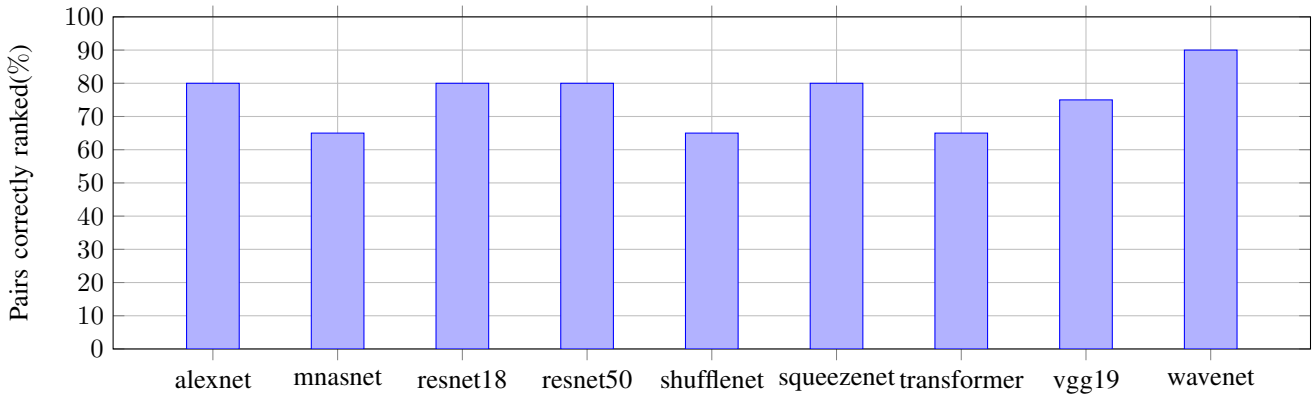

We train our model on a database of $1.6$ million schedules, derived from $10000$ randomly generated ONNX models, converted into Halide pipelines. We use two existing performance models for comparative evaluation. The Halide~\cite{halide_auto} model also uses a neural network, and we train and evaluate it on our train and test set. The TVM~\cite{tvm_model} model does not use a pre-trained model. Instead, it uses adaptive online learning via an exploration phase where different schedule configurations are benchmarked. Since it does not require any pre-training, we used the test split of our dataset on this XGBoost based model.

The plots in figure \ref{error_plots} compare the three performance models. In terms of mean percentage error computed for the test set, our model is $7.75x$ and $12x$ better than the Halide and TVM models, respectively. The graph-based model can learn and represent relationships between different stages, making it more accurate. We also record the maximum error in the test set, and on this metric, our model outperforms the other two by a significant margin. The $R^2$ score, also called coefficient of determination, is a statistical measure of how well a regression model fits the observations. The closer this value is to $1$, the better the fit. The graph-based model achieves a higher $R^2$ score of $0.92$ compared to the Halide and TVM models, which have a score of $0.89$ and $0.81$, respectively.

\subsection{Ranking Performance}

A standard application scenario uses models like ours to rank schedules. A searcher utilizes the ranking information to select a subset of promising schedules and explore them further. In such settings, rather than directly using the actual raw predictions, we use the model to estimate the relative performance of multiple schedules. The more accurate the model, the better it will be at ranking schedules and enable the search to progress in the right direction.

Figure \ref{ranking} displays the pair-wise ranking performance of the model on several hundred schedules for the nine neural networks shown along the x-axis. The schedules were generated using the Halide auto-scheduler. For all possible pair-wise combinations of schedules belonging to a network, we count the number of pairs in which the model assigned a lower run time to the faster schedule. The chart shows the percentage of pairs correctly ranked by the model. The ranking accuracy ranges from $65\%$ for \emph{shufflenet} to as high as $90\%$ for \emph{wavenet}, with an average accuracy of around $75\%$. 

\section{Related work}
\label{relatedwork}
Apart from the Halide~\cite{halide_auto} and TVM~\cite{tvm} models, which we used as baselines for comparative evaluation, there have been several efforts to develop statistical models to predict the performance of deep learning programs. These efforts vary in terms of the core architecture, target environment, feature engineering, etc. As has already been discussed, TVM incorporates two performance models based on \emph{gradient boosted trees} (GBT) and \emph{treeGRUs}, which represent the loop-nest (low-level AST) as a vector embedding and maps the embedding to the cost using a linear layer. The Halide model~\cite{halide_auto} uses a feed-forward neural network. The network learns to predict the run time conditioned on handcrafted features extracted from the program.  Recent work on Halide replaces the feed-forward network with a bi-directional LSTM model and demonstrates significant improvement in prediction accuracy~\cite{value_learn}. The performance model proposed in this work also targets Halide and uses similar handcrafted features. However, our model exploits the inherent graph representation of programs using graph convolutional networks. 

In \cite{tree_lstm}, the authors develop a model to predict the efficacy of code transformations in the Tiramisu compiler. Their composite model built using LSTM and feed-forward networks comprises three distinct modules to embed the computations within a loop, embed the nested loop structure, and a layer to predict the final speed-up. Recently, graph networks based performance models have been proposed for XLA programs on TPUs~\cite{tpu_model}. In addition to using a different graph architecture, their work targets a different class of hardware in contrast to our work which focuses on CPUs. Their approach also decomposes the computational graph of the XLA program into smaller sub-graphs, whereas our model analyzes the entire graph of a Halide pipeline. Moreover, the authors train separate instances of the model for distinct optimization tasks like tile size selection and operator fusion. Our model, on the other hand, is generic and meant for combinatorial optimization search strategies.

The first Halide auto-scheduler~\cite{mullapudi} utilized a simple static analytical model to perform a greedy search in a templatized schedule space. The simplicity of the model led to fast compilation times with no benchmarking or tuning. However, the compiler explored a limited subset of the schedule space and ignored potentially useful schedules comprising multi-level tilings and a wide range of choices for vectorization and parallelism. Subsequent work~\cite{sioutas1} proposed a richer, still manual, model that incorporated the memory access features of modern CPUs like caching and prefetching to optimize the implementation of a Halide operator. An extension to this work~\cite{si2} used the model as a part of a back-tracking search capable of scheduling entire pipelines. In the more general context of applying machine learning to improve compilation frameworks, \cite{si3} provides a comprehensive survey. Ithemal~\cite{ithemal} is a deep learning-based architecture to predict the throughput of straight-line x86\_64 code. 

\section{Future Work}
\label{future}
This work presents preliminary evidence of the benefits of using Graph Networks to model the performance of deep-learning applications. In this section, we discuss several ways to foster better integration and improve the proposed design. 
\subsection{Feature Engineering}
The current model relies upon a very complex feature engineering process. Reliance on manual hand-crafted features limits a model's portability. For example, while the current set of features is applicable across CPU platforms, it would require significant rework when porting to other hardware architectures like GPUs, custom ASICs, and FPGAs. We plan to explore alternative ways to represent deep-learning programs to reduce our reliance on handcrafted features and foster better compatibility across platforms. Such an automated feature engineering strategy would rely on learning representations of the program structure (abstract syntax tree, for example) and learning direct embeddings of tokens in the program. We also intend to study the trade-offs between the two design choices. 
\subsection{Model Improvements and Optimizations}
We evaluated our model on a dataset of $1.6$ million schedules derived from $10,000$ pipelines. Generating this dataset took weeks and required an enormous amount of computing power and resources. We plan to examine the role active learning can play in our application scenario and help retain the same accuracy with a smaller training set. 

As a follow-up to the current work, we are integrating the proposed performance model with the Halide auto-scheduler to gain a measure of scheduling/compilation times and analyze bottlenecks. Using a model like ours for compilation is significantly faster than auto-tuning approaches. However, we plan to compare our model with existing Halide models in terms of compilation times, quality of schedules, etc.

\section{Conclusion}
\label{conclusion}
This paper presented a novel Graph Neural Network-based performance model to estimate the run times of deep learning pipelines implemented using the Halide framework. We proposed a network architecture based on Graph Convolutions capable of capturing interactions between neighboring nodes to generate richer representations that improve prediction performance. The model, implemented in PyTorch, takes features extracted from deep-learning programs implemented in Halide as input and outputs the run time. We generated a dataset of $1.6$ million schedules from around $10,000$ different pipelines and benchmarked them on Intel Xeon CPUs. Experimental evaluation demonstrates that the proposed model outperforms start-of-art Halide and TVM models, with a $7.75x$ and $12x$ reduction in prediction error, respectively. Moreover, our model is approximately $75\%$ accurate when performing a pair-wise ranking of hundreds of schedules derived from $9$ real world pipelines.

\bibliographystyle{IEEEtran}
\bibliography{references}

\end{document}